# Graph Classification and Radiomics Signature for Identification of Tuberculous Meningitis

Snigdha Agarwal, MS[1], Ganaraja V H, MD[2], Neelam Sinha, PhD[3], Abhilasha Indoria, MS[4], Netravathi M, MD[2], Jitender Saini, MD[4*]

[1]Department of Networking and Communication, International Institute of Information Technology, Bangalore, India
[2]Department of Neurology, National Institute of Mental Health and Neuroscience, Bangalore, India
[3]Centre for Brain Research, Indian Institute of Science, Bangalore, India
[4]Department of Neuroimaging and Interventional Radiology, National Institute of Mental Health and Neuroscience, Bangalore, India

## Abstract

**Introduction:** Tuberculous meningitis (TBM) is a serious brain infection caused by Mycobacterium tuberculosis, characterized by inflammation of the meninges covering the brain and spinal cord. Diagnosis often requires invasive lumbar puncture (LP) and cerebrospinal fluid (CSF) analysis.

**Objectives:** This study aims to classify TBM patients using T1-weighted (T1w) non-contrast Magnetic Resonance Imaging (MRI) scans. We hypothesize that specific brain regions, such as the interpeduncular cisterns, bone, and corpus callosum, contain visual markers that can non-invasively distinguish TBM patients from healthy controls. We propose a novel Pixel-array Graphs Classifier (PAG-Classifier) that leverages spatial relationships between neighbouring 3D pixels in a graph-based framework to extract significant features through eigen decomposition. These features are then used to train machine learning classifiers for effective patient classification. We validate our approach using a radiomics-based methodology, classifying TBM patients based on relevant radiomics features.

**Results:** We utilized an internal dataset consisting of 52 scans, 32 from confirmed TBM patients based on mycobacteria detection in CSF, and 20 from healthy individuals. We achieved a 5-fold cross-validated average F1 score of 85.71% for cistern regions with our PAG-Classifier and 92.85% with the radiomics features classifier, surpassing current state-of-the-art benchmarks by 15% and 22%, respectively. However, bone and corpus callosum regions showed poor classification effectiveness, with average F1 scores below 50%.

**Conclusion**: Our study suggests that algorithms like the PAG-Classifier serve as effective tools for non-invasive TBM analysis, particularly by targeting the interpeduncular cistern. Findings indicate that the bone and corpus callosum regions lack distinctive patterns for differentiation.

## 1. Introduction

Tuberculous meningitis (TBM) is a serious infectious disease that affects the membranes surrounding the brain and spinal cord, caused by the bacterium Mycobacterium tuberculosis [1]. This form of meningitis is one of the common extra-pulmonary manifestations of tuberculosis (TB), which can spread from the lungs to the central nervous system, often resulting in life-threatening complications. TBM disproportionately impacts individuals in developing countries, with a higher prevalence in areas with elevated rates of tuberculosis and in populations with compromised immune systems, including young children and individuals with HIV/AIDS [2].

Rapid identification and treatment of TBM are critical for improving patient outcomes and preventing severe neurological damage or death [3]. Early diagnosis is challenging due to the nonspecific symptoms of the disease and the limitations of current diagnostic methods. Currently, two primary methods are employed to diagnose TBM: 1) a diagnostic approach using non-invasive techniques, which utilizes contrast-enhanced imaging, and 2) Biomarker Analysis, which involves performing an invasive lumbar puncture (LP) to obtain cerebrospinal fluid (CSF) for the detection of cell count, protein, glucose and special stains and culture of the CSF. Both techniques are commonly used; however, LP, being an invasive procedure, requires precision and expertise. It is not feasible in all situations due to complications such as hydrocephalus or other associated contraindications. The use of gadolinium-based contrast agents in Magnetic Resonance Imaging (MRI) techniques carries a rare but serious risk of long-term accumulation in the brain and other tissues. These risks highlight the necessity for alternative diagnostic methods that reduce reliance on invasive procedures and gadolinium use for patients requiring multiple scans for chronic conditions like TBM.

To address these concerns, we attempt to identify TBM patients using advancements in Machine Learning (ML) and Artificial Intelligence (AI). This study discusses approaches for diagnosing TBM using non-contrast T1-weighted (T1w) MRI scans using radiomics [4] and a proposed graph-based method using feature extraction from image patches of relevant brain regions. The block diagram in Figure 1 captures the overview of the proposed methodology.

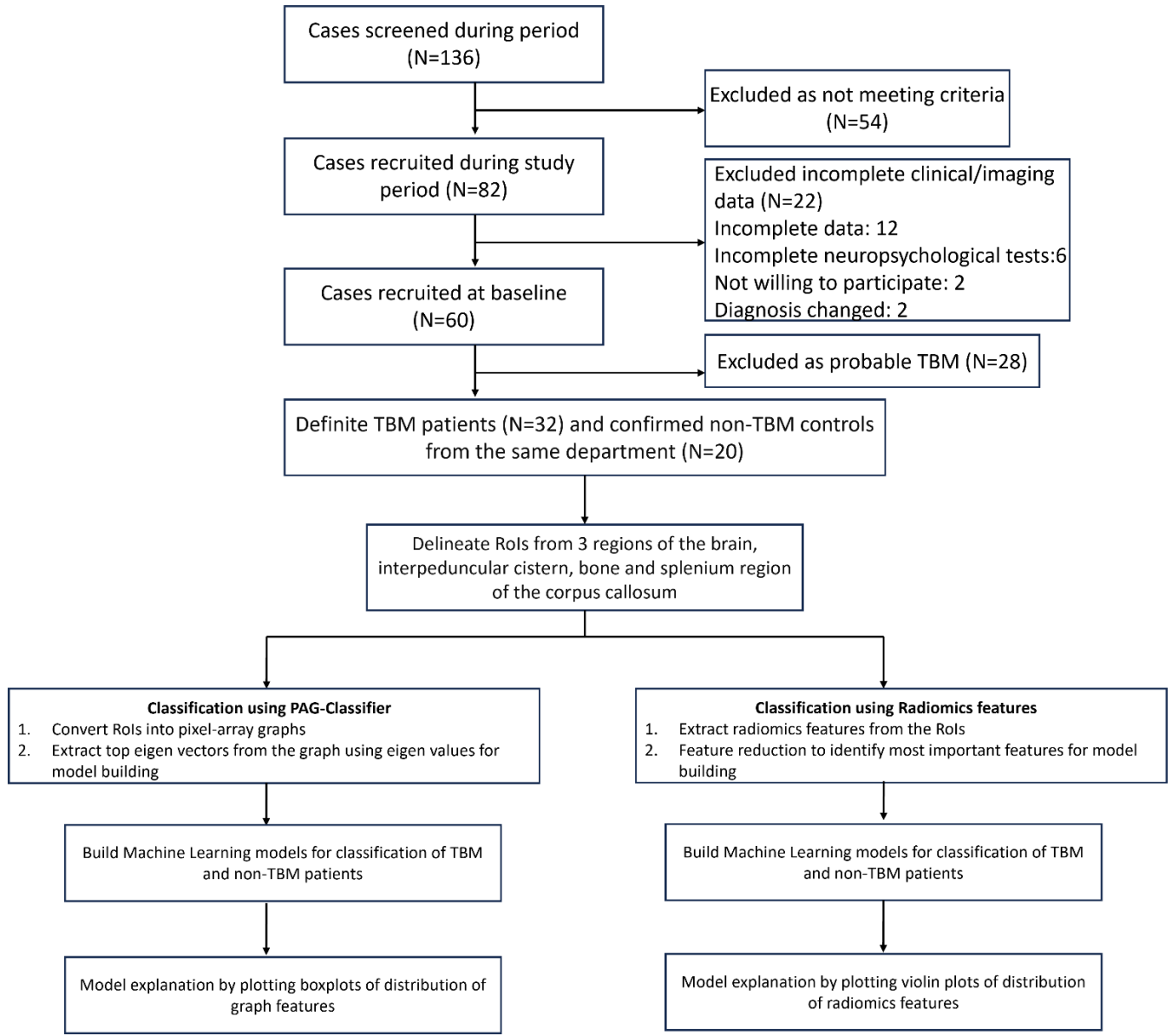

**Figure 1**: *Block diagram showing the overall workflow of our proposed methodology.*

## 2. Materials and Methods

### 2.1. Subject Recruitment, Clinical Evaluation and Imaging

A total of 60 adult patients with clinically suspected Tuberculous meningitis and 20 healthy controls were recruited from the outpatient and emergency services of the neurology department at a tertiary university hospital. To confirm tubercular infection, CSF analysis was performed using tests such as Ziehl Neelsen smear, Gene Xpert, Mycobacterial Growth Indicator Tube (MGIT), and culture by Lowenstein Jensen medium. Out of these, 32 patients who showed evidence of mycobacteria through these tests were confirmed as definite TBM patients and included in our study.

MRI images for each participant were acquired on a 3T SIEMENS MAGNETOM Skyra Syngo scanner. The imaging protocol included a 3D T1w Magnetization Prepared Rapid Gradient Echo (MPRAGE) sequence, covering the entire brain. The parameters for the scan were as follows: repetition time (TR) = 8.1 ms, echo time (TE) = 3.7 ms, flip angle = 8°, sense factor = 3.5, field of view (FOV) = 256 mm × 256 mm × 155 mm, voxel size = 1 mm × 1 mm × 1 mm, slice thickness = 1 mm, and acquisition matrix = 256 × 256. The preference for T1w scans over Fluid-attenuated inversion recovery (FLAIR) or T2-weighted (T2w) scans is due to the significant imaging artefacts associated with the movement of CSF observed in FLAIR sequences. Additionally, T2w sequences frequently exhibit the flow void phenomenon, attributed to the turbulence of flowing CSF. Although T2w 3D scans would have been ideal for our analysis, logistical constraints prevented us from performing 3D T2w scans for the TBM patients.

| Region of Interest | Description |
|---|---|
| 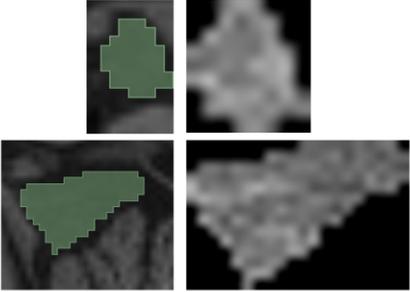 | • We extract two separate 3D regions of interpedencular cisterns from the full MRI volume.<br>• The beginning of the annotation is from a slice where the cisterns appear with maximum clarity.<br>• The two regions in some cases can be from two different slices. |
| 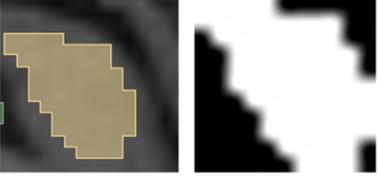 | • We extract 3D region of the bone similarly from the slice selected for interpedencular cistern. In case of multiple slices, any of the two slices are used. |
| 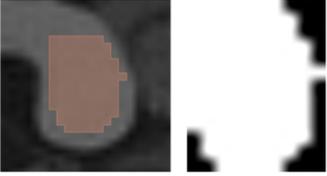 | • A 3D ROI is extracted from the Splenium region of the corpus callosum from the previously selected slices |

**Figure 2**: *Description of the methodology for selecting ROIs and sample annotated ROIs from the bone, corpus callosum, and interpeduncular cistern regions.*

### 2.2. *Annotation of Regions of Interest*

To assess if TBM can be successfully diagnosed from MRI scans, our initial step involves identifying Regions of Interest (ROIs) within three distinct areas of the brain scan. To annotate the RoIs, we use the segmentation tool in 3DSlicer[1]. The schematic diagram of our methodology to delineate ROIs is present in Figure 2.

### 2.3. *Study Objectives*

For classification of TBM patients, the objectives are as follows:

1) Use 3D gray-scale volume ROIs of bone, corpus callosum and interpeduncular cisterns of both patients and controls, $I(x, y, z)$ where $x, y, z$ represent the spatial dimensions of the image.

---

[1] https://www.slicer.org/

2) Learn an optimal classification function $f$ such that $f: I(x, y, z) \rightarrow \{0,1\}$, where $0$ signifies controls and $1$ signifies patients using our proposed Pixel-arrays Graphs (PAG-Classifier) and Radiomics feature-based classifier.

*2.4. Classification based on Pixel-array Graphs*

We introduce a novel method for classifying TBM patients through the application of Graph Theory [7]. By employing this strategy, we construct mathematical models known as graphs to represent pairwise relationships among arrays of pixels from the ROIs. We construct separate graphs for each image patch in the bone, corpus callosum, and the left and right cistern regions, resulting in a total of four distinct regions.

**Graph Creation:** We conceptualize each image $I(x, y, z)$ normalized as $I(x, y, z) = I(x, y, z)/max(I(x, y, z))$ as a unified pixel-array graph with each array in $z$ dimension functioning as a node within this graph. The edges between these nodes encapsulate the spatial relationship among the pixel-arrays. Mathematically, our graph can be denoted as $G = (V, E)$, where $V$ signifies the total number of nodes calculated as $x \times y$ and $E$ is the set of edges connecting the nodes. Each edge $(U, V) \in E$ represents a connection between pixel-arrays $U$ and $V$. This transformation converts the 3D MRI scan to a 2D graph, where $V$ (the total nodes) are the total pixels in each x-y plane. The schematic diagram of our methodology of graph creation is present in Figure 3a. The weight of the connection is calculated using a Mutual Information Score $M(U, V)$ defined as follows,

$$M(U, V) = \sum_{u \in U} \sum_{v \in V} p(u, v) \log \left( \frac{p(u,v)}{p(u)p(v)} \right) \qquad (1)$$

where, $p(u, v)$ is the joint probability distribution function and $p(u)$ and $p(v)$ are the marginal probability distribution functions of $U$ and $V$. The edge weights are further normalized as below,

$$M_{norm}(U, V) = \frac{M(U,V) - M_{min}}{M_{max} - M_{min}} \quad (2)$$

The edges between pixel-arrays are un-directed. We remove edges with $M_{norm}(U, V)$ less than $0.5$ creating a connected graph with absence of a relationship between certain pixel-array pairs. Sample pixel-array graphs for all the regions are present in Figure 3b.

**Feature Extraction:** To extract features from the created graphs for classification, we convert the graphs into an adjacency matrix. For a graph $G = (V, E)$, the adjacency matrix $A$ is defined as follows: Let $A$ be $n \times n$ matrix, where $n$ is the number of nodes in $V$, such that as each element $a_{ij}$ of $A$ is given by:

$$a_{ij} = \begin{cases} M_{ij}, & \text{if there is a node } i \text{ from node } j \text{ with weight } M_{ij} \\ 0, & \text{otherwise} \end{cases}$$

Since our graphs are un-directed, the adjacency matrix is symmetric, i.e., $a_{ij} = a_{ji}$. We further extract the eigenvalues and eigenvectors from the adjacency matrix, represented as,

$$A\vec{v_i} = \lambda_i \vec{v_i}, \text{for } i = 1, 2, \ldots, n$$

Here, $\lambda_i$ signifies the eigenvalues of matrix $A$, and $\vec{v_i}$ are the corresponding eigenvectors. In this context, $\vec{v_i}$ is a non-zero vector, and $\lambda_i$ is a scalar that indicates the factor by which the eigenvector is scaled during the linear transformation described by $A$. Eigenvalues and eigenvectors are pivotal in understanding the structure of the graph, such as its connectivity and the presence of clusters, which are instrumental for extracting features for classification tasks. We sort the eigenvectors in descending order of their eigenvalues, $\lambda_1 > \lambda_2 > \cdots > \lambda_n$ and select the top 8 eigenvectors for further analysis. These vectors are then flattened to construct features for every graph corresponding to each region. Subsequently, these features are utilized to train and validate the classification model.

**Graph Classification Model Building:** For classification we build separate models for left cistern, right cistern, bone and corpus callosum regions. he combined predictions for the left and right cisterns use AND logic, formulated as follows,

$$F_{fc}: I(x, y, z) \to \{0,1\} \wedge F_{sc}: I(x, y, z) \to \{0,1\} \tag{3}$$

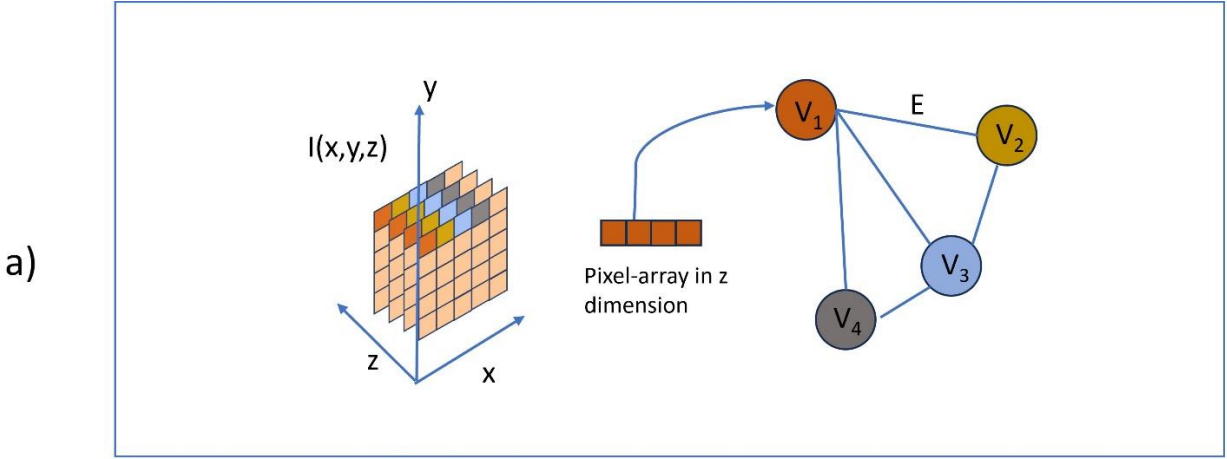

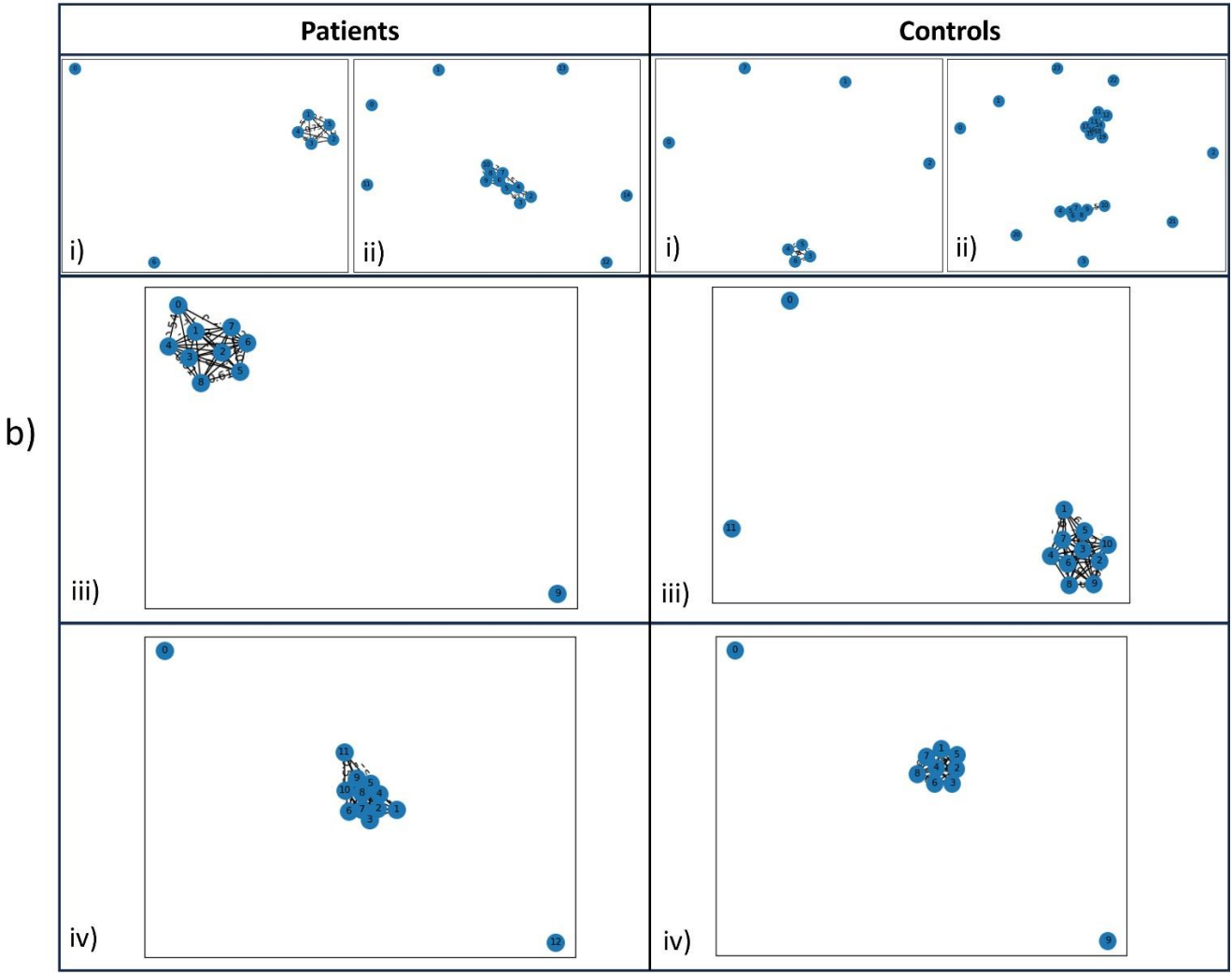

***Figure 3:*** *a) Schematic diagram of graph creation from ROIs, b) shows sample generated graphs for, i) the left cistern, ii) right cistern, iii) bone, iv) corpus callosum ROIs for both patient and control groups.*

where the TBM incidence is affirmed only if both models concurrently indicate incidence. In our investigation, we employ state-of-the-art (SOTA) algorithms, namely, the Random Forests Classifier [8], Support Vector Machines (SVM) [9], and LightGBM Classifier [10], to develop our model. The training and evaluation of the model are conducted using a 5-fold cross-validation approach, which incorporates both random shuffling and label stratification for robustness. To optimize model performance, we undertake hyperparameter optimization through the Grid Search [11] method. Among the evaluated models, we identify and select the one with superior performance. The efficacy of our chosen model is quantified using metrics such as average accuracy, F1 score, and Area Under the Receiver Operating Characteristic (AUROC) [14] over the cross-validation process.

### 2.5. Classification based on Radiomics Features

In this method, we build a radiomics signature using a Gradient Boosted Machine (GBM) model to classify patients and controls from MRI scans.

**Feature Extraction:** We extract a feature vector of length $n$ from all the four ROIs of bone, corpus callosum and interpeduncular cisterns from MRI scans of patients and controls using the function $g: I(x, y, z) \rightarrow R^n$. The resulting feature vector serves as input to the classification function $f$, mathematically formalized as $f\left(g\big(I(x, y, z)\big)\right) \rightarrow \{0,1\}$. We build separate models for bone, corpus callosum and interpeduncular cistern ROIs. Before feature extraction, the 3D images are z-score normalized, using the normalization function $N$ defined as,

$$N(I(x, y, z) = (I(x, y, z) - \mu_I)/\sigma_I \qquad (4)$$

We use the sitkBSpline [12] interpolation technique for voxel resampling. We extract the following features from the images, 1) shape, 2) first-order, 3) glcm, 4) glrlm, 5) glszm, 6) gldm, 7) ngtdm on the original images as well as applying the following image filters, 1) exponential, 2) gradient, 3) LBP2D, 4) LBP3D, 5) LoG, 6) logarithm, 7) square, 8) squareRoot, 9) wavelet (exhaustive list of all combinations of applying either a High or a Low pass filter in each of the three dimensions). This results in a total of 1713 extracted features.

**Feature Reduction:** Given that the total number of scans is significantly lower than the number of extracted features, we employ feature reduction techniques to reduce data dimensionality. To prevent overfitting due to multicollinear features and to enhance model explainability, we use the Pearson correlation coefficient [13] to eliminate correlated features. Specifically, we remove features with less than 1% correlation with the target variable and those with more than 95% correlation with another feature. This process results in a total of 197 features. We further use a Gradient Boosting model (LightGBM) [10] to identify the top features with a split importance of >= 1 using model feature importance. The selected features for each of the three regions are shown in Figure 4.

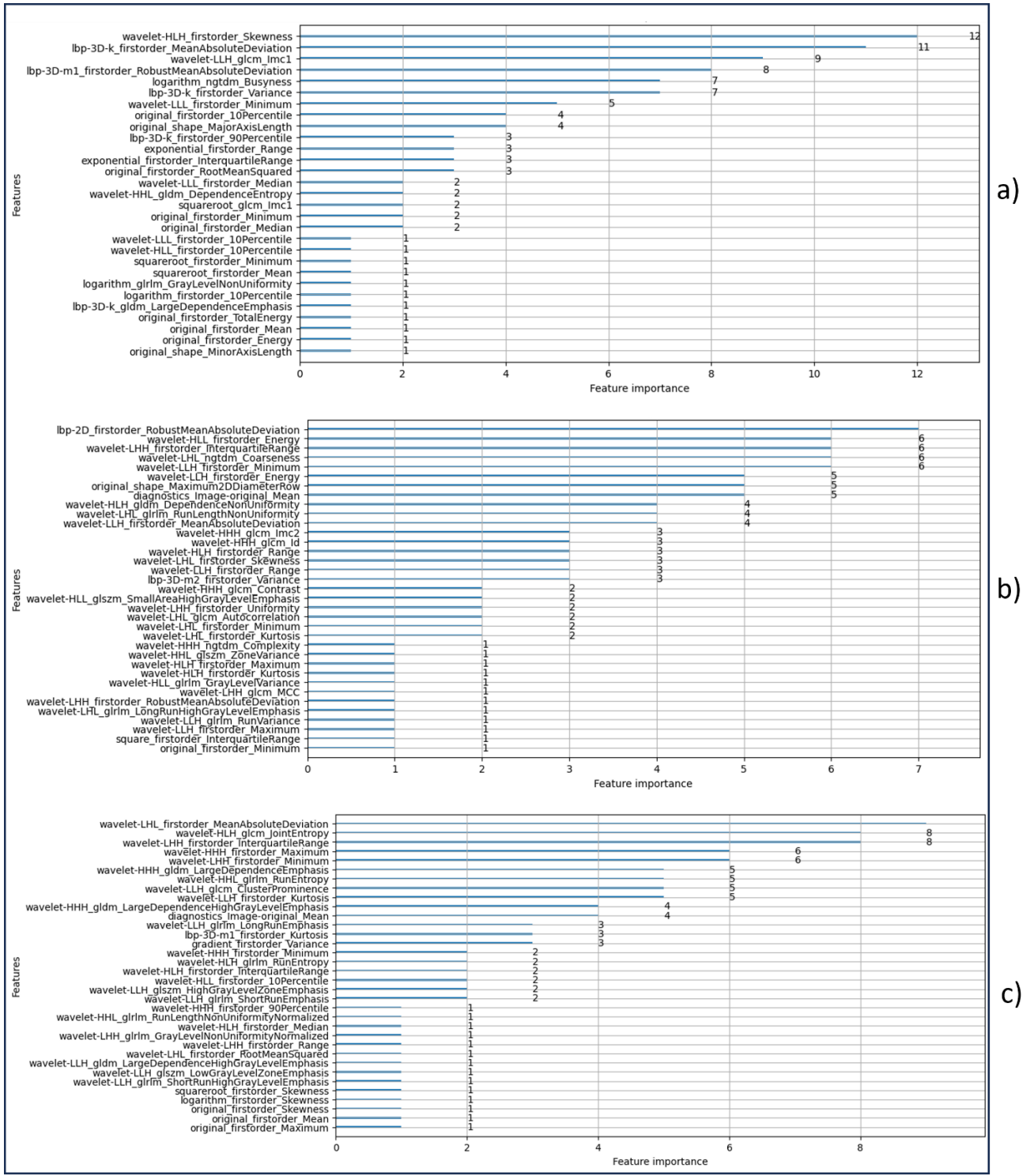

**Figure 4**: *a) Shows the selected features for classifying patients and controls using interpeduncular cistern ROIs, b) shows the selected for the classification model using bone ROIs and c) shows the selected features for the classification model using corpus callosum ROIs. The selection of features is based on the frequency of their usage in candidate models.*

**Classification Model Building:** We build another LightGBM model to classify TBM patients from controls. We allocate 80% of the dataset for training and the remaining 20% for testing, employing stratified sampling and random shuffling for selection. To assess model performance, a 5-fold cross-validation strategy is utilized. We build the model using a learning rate of 0.01, a total of 1000 boosted trees, and 31 leaves with the F1 score serving as the optimization criterion. We build separate models for each of the three regions under study. The performance metrics reported include the average accuracy, average F1 score, and average AUROC. An in-depth discussion of the results is provided in the subsequent section.

Table 1: Comparison of evaluation metrics for each region using PAG-Classifier.

| Region | Algorithm | Accuracy (avg %) | F1 score (avg %) | AUROC (avg %) |
|---|---|---|---|---|
| Interpeduncular cisterns | Random Forest | 90.91 | 85.71 | 93.75 |
| | SVM | 64.55 | 65.41 | 69.95 |
| | Light GBM | 74.73 | 69.54 | 77.73 |
| Bone | Random Forest | 65.45 | 44.76 | 59.38 |
| | SVM | 57.45 | 43.43 | 54.31 |
| | Light GBM | 65.09 | 52.42 | 61.37 |
| Corpus Callosum | Random Forest | 57.45 | 32.38 | 55.79 |
| | SVM | 50.36 | 24.44 | 45.96 |
| | Light GBM | 58.91 | 47.33 | 56.05 |

## 3. Results

Using our proposed approach of PAG-Classifier, we can correctly differentiate between TBM patients and controls using the interpeduncular cistern region with an average accuracy of 90.91% and average F1 score of 85.71% using the Random Forest algorithm. The detailed results are present in Table 1. To correlate the model performance with the features of the graphs constructed using this region, we run an explainability analysis [15]. We summarize the characteristics of the graphs using the following features,

- Number of edges: Total number of edges in the graph

- Number of nodes: Total number of nodes in the graph

- Edge weight: Average weight of the edges in the graph calculated as,

$$edgeweight_{avg} = \frac{1}{|E|} \sum_{(U,V) \in E} M_{\text{norm}}(U,V)$$

where, E is the set of all edges in the graph, $|E|$ denotes the number of edges, and $(U,V)$ represents each edge connecting the pixel-arrays $U$ and $V$. The function $M_{\text{norm}}$ as described in (2), is applied to each edge to determine its weight.

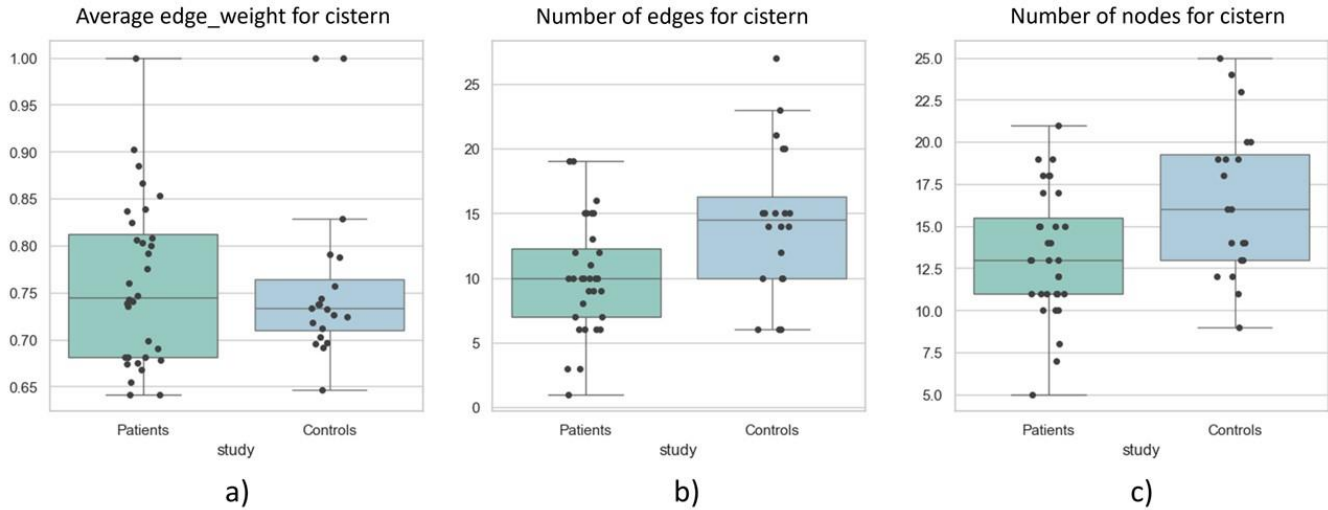

a)                     b)                     c)

***Figure 5:*** *Boxplot comparison highlighting differences in the distribution of graph features between patient and control groups within the most discriminative region, the interpeduncular cistern. a) illustrates a broader spread in the distribution of average edge-weight for patients compared to a narrower range for controls. b) depicts that the distribution of the number of edges varies between the two groups, with c) demonstrating a notable disparity in the number of nodes, which are significantly greater for controls than for patients, indicating potential biomarkers for TBM detection.*

We utilize boxplots to examine the variability in these features across the generated graphs for both patients and controls. In Figure 5., we observe that the distribution of the features of the graphs differ significantly between the patient and control groups.

Table 2: Comparison of evaluation metrics for each region using radiomics features.

| Region | Accuracy (avg %) | F1 score (avg %) | AUROC (avg %) |
|---|---|---|---|
| Interpeduncular cisterns | 92.18 | 92.85 | 93.08 |
| Bone | 78.36 | 80.65 | 76.90 |
| Corpus Callosum | 70.18 | 71.19 | 70 |

Using the radiomics approach, we see that our model can correctly differentiate between TBM patients vs controls using the interpeduncular cistern region with an F1 score of 92.85% and an accuracy of 92.18%. The average AUROC score is 93.08% as summarized in Table 2. To mitigate a black box effect of our model and to bring in explainability to the model performance, we use Shapley Additive Explanations (SHAP) [16]. From Figure 6a, we see that features like "lbp-3D-k-firstorder-MeanAbsoluteDeviation", "wavelet-HHL-glrlmShortRunLowGrayLevelEmphasis", "waveletLLHfirstorderMeanAbsoluteDeviation" significantly enhance the model's learning capabilities. To validate these findings with empirical data, Figure 6b presents a comparative analysis of the distributions of these features using violin plots [17]. This analysis reveals distinct, non-overlapping ranges of variability for these features between patients and controls, showing their diagnostic relevance.

From both our approaches, the graph-based and radiomics, we see that our model is not able to achieve a good performance on the bone and corpus callosum regions. The average of evaluation metrics for the graph-based model is less than 60% and an average of 75% for the radiomics approach. These results indicate that the bone and corpus callosum regions lack distinct patterns that are effective for differentiating TBM patients from controls.

In Table 3, we compare the results of our best performing models from the PAG-Classifier model and the radiomics model with the current SOTA model.

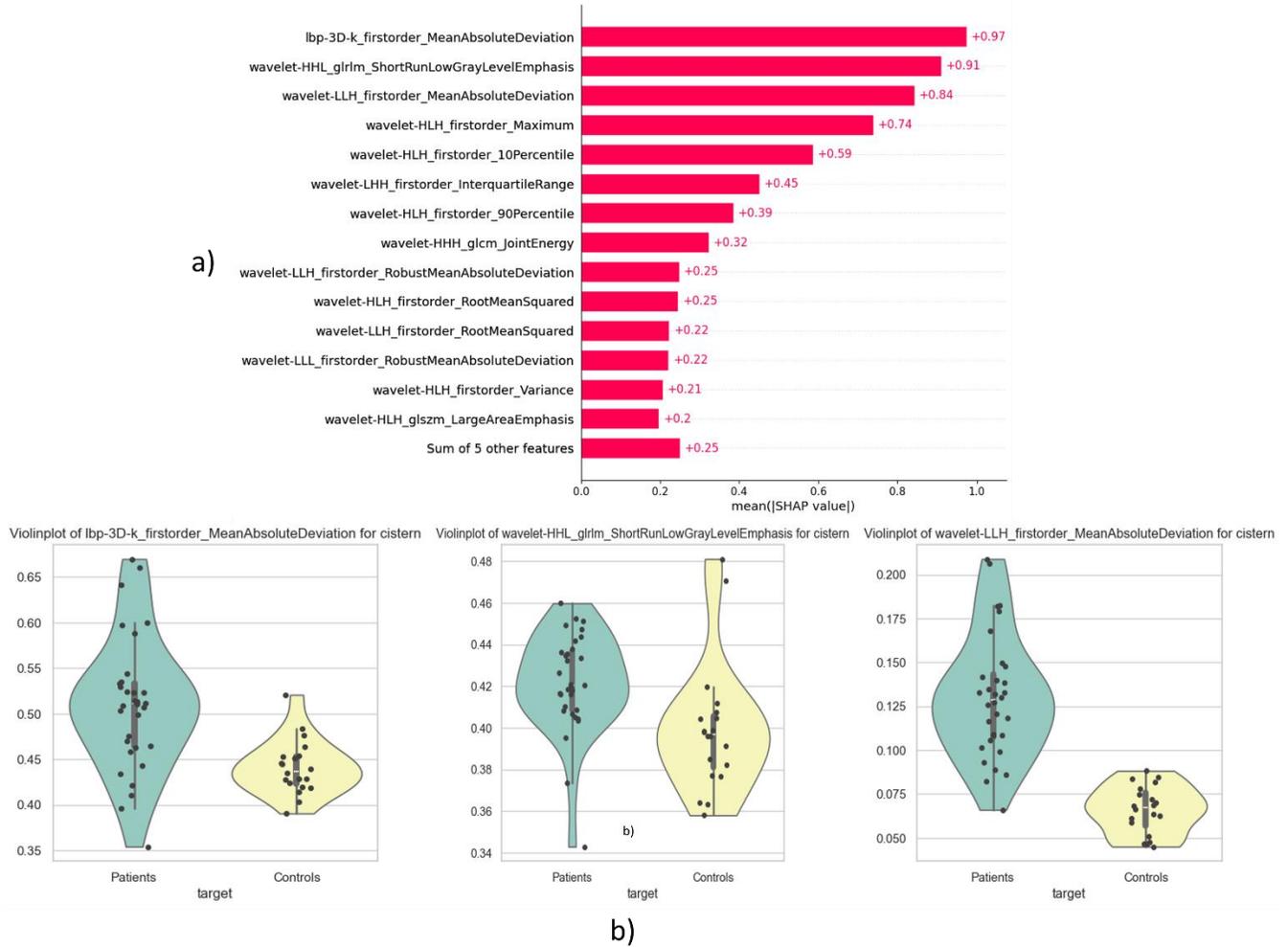

***Figure 6:*** *a) Presents SHAP value analysis, showing that the features "lbp-3D-k-firstorder-MeanAbsoluteDeviation," "wavelet-HHL-glrlm-ShortRunLowGrayLevelEmphasis," and "wavelet-LLH-firstorder-MeanAbsoluteDeviation" contribute the most to model performance. b) shows violin plots that confirm the distinct distributions of these key features between patient and control groups, validating their efficacy in training the model.*

Table 3: Comparison of evaluation metrics with the SOTA

| Approach | Algorithm | Accuracy (avg %) | F1 score (avg %) | AUROC (avg %) |
|---|---|---|---|---|
| Radiomics [8] | SVM | 0.741 | - | 0.751 |
| **PAG-Classifier (Ours)** | Random Forest | 90.91 | 85.71 | 93.75 |
| **Radiomics (Ours)** | LightGBM | 92.18 | 92.85 | 93.08 |

## 4. Discussion

In this study, we develop a diagnostic algorithm for TBM using non-invasive methods, employing AI-ML models applied to radiomics. Radiomics can help to improve the diagnostic accuracy and assist in decision making based on large number of medical image features. The classification results from our proposed methods and algorithms demonstrate their potential as effective tools for non-invasive TBM diagnosis. Our study also suggests that cisternal space in the basal region effectively differentiates TBM from healthy controls. Previous studies on diagnosing TBM have focused on classification based on demographic and clinical characteristics of patients, as well as protein levels in CSF using proteomics [18, 19, 20]. Though these diagnostic models have tried to diagnose TBM, they all require invasive LP technique for CSF extraction. Few studies have attempted to differentiate TBM from non-TBM patients such as those with Bacterial Meningitis (BM) and Viral Meningitis (VM), using demographic features and clinical symptoms. These studies used Logistic Regression and Decision Tree algorithms [22, 23, 24]. In a study by Dendane et al. [22], various clinical and laboratory features of TBM and BM were utilized for classification. They use a diagnostic scoring system developed with a multivariate logistic regression model. This model included clinical and CSF parameters, without taking any imaging characteristics into consideration. In another study [23], a clinical decision support system was created using a large database of meningitis patients. This system was based on interpretable tree-based machine learning models and knowledge-engineering techniques, aiming to assist clinicians in diagnosing meningitis and supporting them in various stages of patient care. However, this study was mainly focused on diagnosis of meningitis rather than identifying different subtypes of meningitis such as TBM. In a study by Joeng et al. [24], machine learning techniques were used for the diagnosis of TBM and VM. The study involved comparing various clinical and CSF parameters of TBM with those of VM. An Artificial Neural Network model was created, which was found to have better diagnostic performance than non-expert clinicians. However, a major limitation was that not all TBM patients in the cohort were definite TBM with 21 out of 60 classified as probable TBM. Additionally, imaging features were not taken into consideration in this study.

Currently, the only radiomics study integrates T2w and FLAIR imaging signatures with deep learning for segmenting and identifying subtle changes in the basal cisterns, aiding in the diagnosis of TBM. In this study, SVM and Logistic Regression classifiers were used for identification [25]. Building on this initial research, our study utilizes T1w scans, addressing the limitations of using T2w and FLAIR MRI scans for diagnosing TBM [25]. Notably, our study found that T1w scans are more sensitive than T2w and FLAIR for detecting TBM, contrary to the findings reported in previous study. However, the effectiveness of image segmentation on T1w scan compared to T2w and FLAIR depends on various factors, including the specific characteristics of the TB meningitis lesions, the imaging protocols used, and the radiomics analysis methods used. T2w imaging, sensitive to changes in tissue water content, may not consistently provide optimal contrast for highlighting subtle meningeal abnormalities or differentiating them from other brain structures. FLAIR imaging, which is highly sensitive to CSF, excels at detecting brain parenchymal abnormalities like white matter lesions but might not always yield the best contrast for detecting meningeal abnormalities associated with TB meningitis [25, 26, 27].

Differences in T1w and T2w scan analyses might also stem from changes in local CSF fluid dynamics at the basal cisterns due to the inflammatory processes of TBM, which can alter T2w and FLAIR signals [25]. Hence, T1w images, which are minimally affected by exudative protein-associated fluid dynamics, are deemed more appropriate for constructing signatures by extracting radiomics features. We also utilized the bone region in both TBM patients and controls for internal validation to enhance differentiation. The pathology of TBM is primarily based on the inflammatory response mediated through various inflammatory markers, which leads to exudation, especially prominent in prepontine regions [28]. This could be one of the reasons for better identification of TBM pathology in basal cisternal region.

Strengths of our study include using a definitive cohort of TBM patients, all confirmed to have mycobacteria in their CSF, compared to healthy controls. This study pioneers a radiomics approach for diagnosis of TBM and highlights the potential of non-invasive diagnostic techniques. This is also the first study to identify TBM patients using non-contrast T1w scans without directly observing basal exudates typically seen in TBM, proving effective in TBM detection. However, the study's major limitation is its non-blinded nature and small sample size, necessitating further research with larger cohort. Inclusion of other non-tubercular meningitis, such as pyogenic meningitis, could have helped in improving the diagnostic system.

Our study introduces a novel method that facilitates the early identification of TBM patients, particularly valuable since conventional diagnostic methods often involve delays in performing LP due to the clinical status of patients and associated contraindications. While invasive procedures such as LP remain important in certain scenarios for definitive diagnosis, non-invasive techniques based on this radiomic signatures serve as valuable adjuncts in the diagnostic workup of TB meningitis for early diagnosis. These techniques offer numerous advantages in terms of safety, patient acceptability, and accessibility. This is an initial study and further research with larger samples is necessary. We also introduce an innovative approach using Pixel-array Graphs to classify TBM patients, aiming to enhance the accuracy and reliability of TBM detection.

## 5. Conclusion

Our study highlights the effectiveness of non-invasive TBM diagnostic tools, specifically demonstrating the utility of the Pixel-array Graphs (PAG) classifier in analysing the interpeduncular cistern. Our findings show the limitations of utilizing the bone and corpus callosum regions for TBM differentiation due to their lack of distinctive patterns. Additionally, the implementation of SHAP values and metadata analysis has significantly improved the interpretability of our models, facilitating a deeper understanding of their decision-making processes. This enhancement aids in refining diagnostic accuracy and offers valuable insights into the key factors influencing model predictions, thus contributing to the advancement of non-invasive methods for TBM diagnosis. Overall, our results show that our suggested approaches and algorithms could serve as effective tools for conducting non-invasive TBM analyses.

## 6.    Data and Code Availability

The raw data supporting the conclusions of this article will be made available by the authors, without undue reservation, to any qualified researcher.